\pdfoutput=1
\documentclass[11pt]{article}
\usepackage[dvipsnames]{xcolor}
\usepackage{ACL2023}
\usepackage{times}
\usepackage{latexsym}
\usepackage{amsmath}
\usepackage{graphicx}
\usepackage{booktabs}
\usepackage[T1]{fontenc}
\usepackage[utf8]{inputenc}

\usepackage{microtype}

\usepackage{inconsolata}

\title{Personalized Abstractive Summarization by Tri-agent Generation Pipeline}
\author{\textbf{Wen Xiao}\footnotemark[2]\footnotemark[3]\hspace{.3em}\qquad Yujia Xie\footnotemark[3]\hspace{1.9em} \\
\textbf{Giuseppe Carenini}\footnotemark[2]\hspace{1.9em}\textbf{Pengcheng He}\footnotemark[3]\hspace{1.9em}\vspace{4pt}\\
\footnotemark[2]\hspace{.4em}University of British Columbia, Vancouver, Canada 
\\ \footnotemark[3]\hspace{.4em} Microsoft Azure AI\\
\texttt{\small{\{carenini\}@cs.ubc.ca}},\\
\texttt{\small{\{wxiao,yujiaxie,penhe\}@microsoft.com
}}}

\begin{document}
\maketitle
\begin{abstract}

% Tailoring outputs of large language models, such as ChatGPT, to implicit user preferences remains a challenge despite their impressive generation quality. In this paper, we propose a tri-agent generation pipeline consisting of a generator, an instructor, and an editor to enhance the personalization of generated outputs. The generator produces an initial output, the trained instructor automatically generates editing instructions based on the user's preference , and the editor generates a revised output aligned with user preferences. The inference-only large language model (ChatGPT) serves as both the generator and the editor, while a smaller model acts as the instructor to guide the generation process towards users' preference. The instructor is trained using editor-steered reinforcement learning, leveraging feedback from the large-scale editor model  to optimize instruction generation. Experimental results on two abstractive summarization datasets demonstrate the effectiveness of our approach in generating outputs that better fulfill user expectations.
Tailoring outputs from large language models, like ChatGPT, to implicit user preferences remains a challenge despite their impressive generative capabilities. In this paper, we propose a tri-agent generation pipeline comprising a generator, an instructor, and an editor to enhance output personalization. The generator produces an initial output, the instructor automatically generates editing instructions based on user preferences, and the editor refines the output to align with those preferences. The inference-only large language model (ChatGPT) serves as both the generator and editor, with a smaller model acting as the instructor to guide output generation. We train the instructor using editor-steered reinforcement learning, leveraging feedback from a large-scale editor model to optimize instruction generation. Experimental results on two abstractive summarization datasets demonstrate the effectiveness of our approach in generating outputs that better meet user expectations.
\footnote{Code is available at \url{https://github.com/Wendy-Xiao/chatgpt_editing_summ}}

\end{abstract}

\section{Introduction}
Large language models, exemplified by prominent models such as InstructGPT~\cite{instructgpt} and ChatGPT\footnote{\url{https://openai.com/blog/chatgpt}}, have emerged as essential resources in the field of natural language processing (NLP). These models have shown an extraordinary level of proficiency across a broad spectrum of NLP tasks, including machine translation, question answering, and text summarization. In light of their potential to drive further innovation in language-based technologies, the research community has exhibited growing enthusiasm for exploring and advancing large language models.
\begin{figure}
    \centering
    \includegraphics[width=\linewidth]{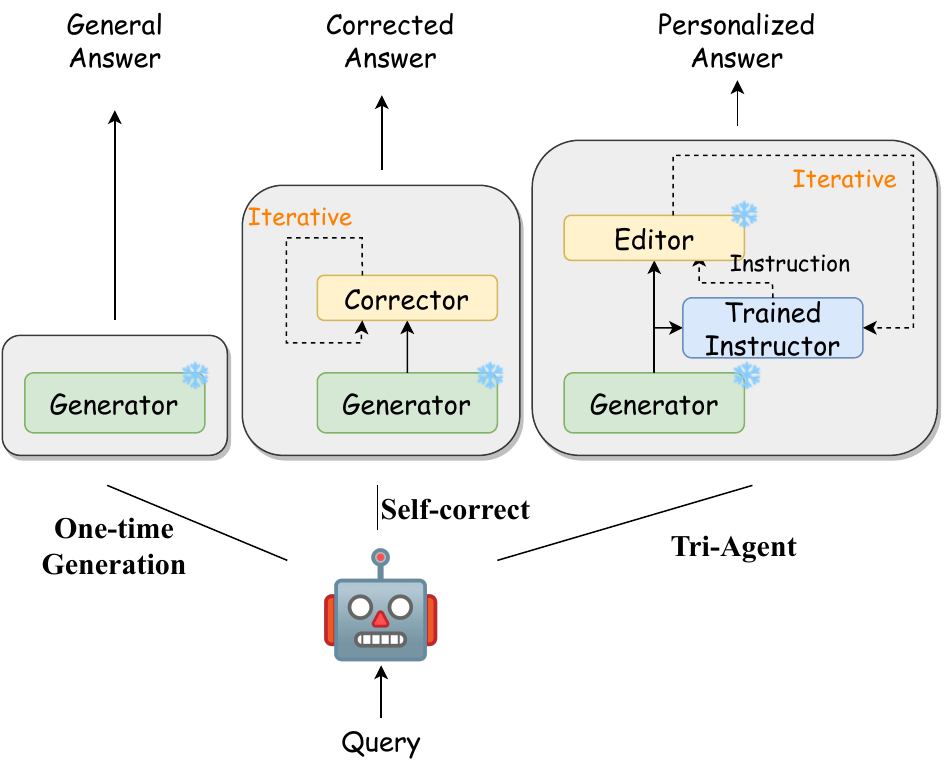}
    \caption{Comparison between different generation paradigms. The left one is the general one-time generation process, the middle one is from \citet{self_correct}, which uses a trained corrector to make corrections on the generated text, usually dealing with specific issues, like eliminating hallucination or toxicity,  and the right one is the proposed tri-agent pipeline.
    % The Generator and Editor in the tri-agent pipeline can be inference-only large language models.
    % The Generator and Editor are frozen large language models in the proposed tri-agent pipeline. 
    % \yujia{Should we highlight it with which parts are frozen?}\wen{added, please check}
    }
    \label{fig:intro}
\end{figure}
% However, despite the impressive generation quality achieved by these models, a persistent challenge lies in tailoring their outputs to meet user's specific needs~\cite{liu2022improving}. In several scenarios, it has been observed that the outputs of language models do not consistently satisfy users' requirements or expectations~\cite{bubeck2023sparks}. A prevalent approach to addressing this limitation involves the careful crafting of prompts to steer the models in producing outputs that better align with users' objectives. Nonetheless, as noted in existing research~\cite{reid-neubig-2022-learning}, the conventional one-time left-to-right generation process of language models contrasts with the iterative refinement and editing approach commonly employed by humans. Furthermore, prior works~\cite{NEURIPS2019_675f9820,reid-zhong-2021-lewis} have demonstrated the efficacy of the generate-and-edit process compared to one-time generation, even with a single editing iteration. Motivated by these findings, this paper explores the integration of large language models (ChatGPT) into an iterative editing pipeline. 
\begin{figure*}
    \centering
    \includegraphics[width=\linewidth]{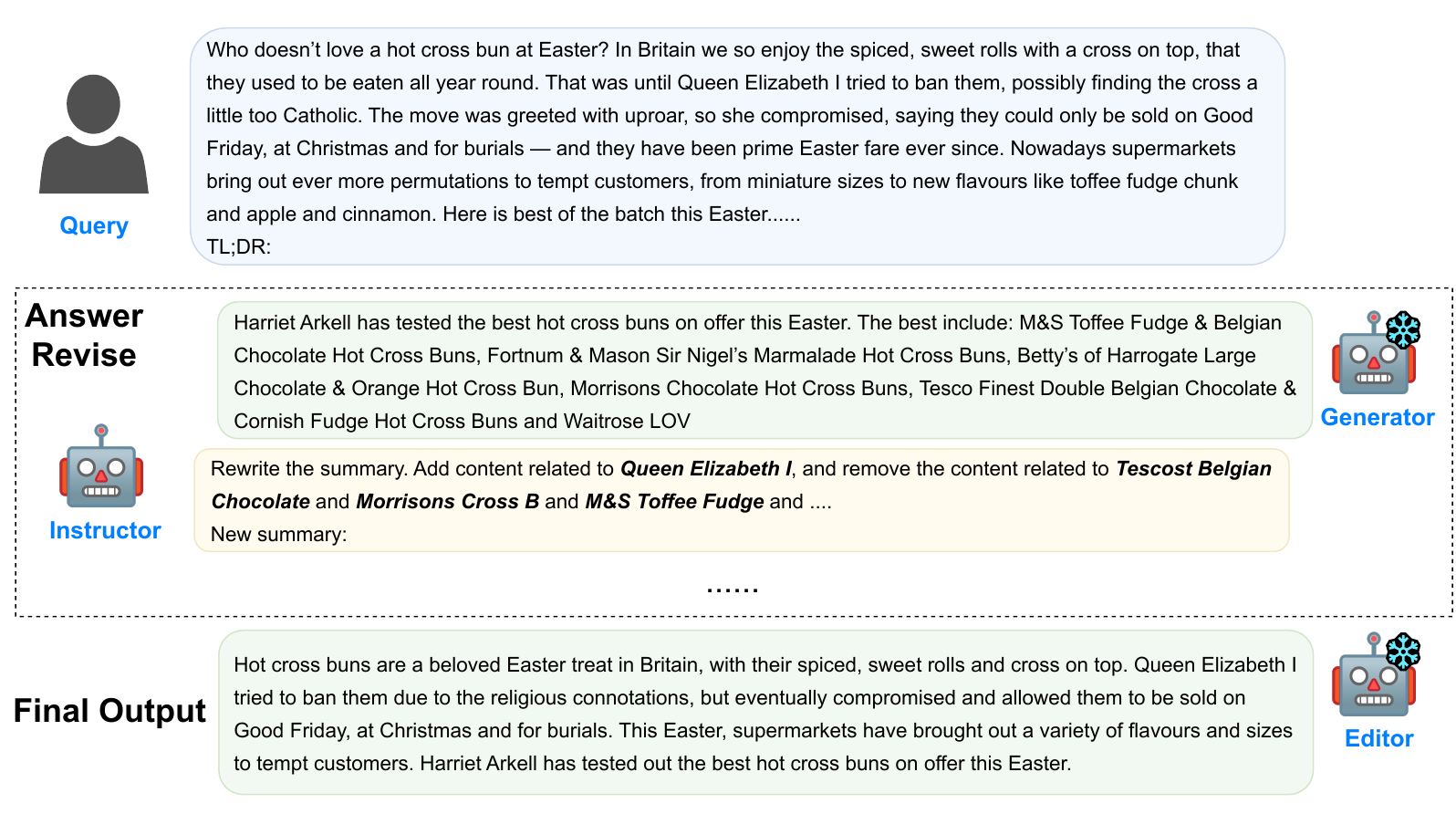}
    \caption{An illustration of the proposed tri-agent generation pipeline. When a query is given, the generator first generates an initial answer, and the instructor provide an instruction on how to make the answer more tailored to user's preference, and finally the editor generates a personalized answer with the given instruction.}
    \label{fig:pipeline}
\end{figure*}
However, despite the impressive generation quality achieved by these models, a persistent challenge lies in tailoring their outputs to meet user's preference~\cite{liu2022improving}. In several scenarios, it has been observed that the outputs of language models do not consistently satisfy users' preferences or expectations~\cite{bubeck2023sparks}. A prevalent approach to addressing this limitation involves the careful crafting of prompts to steer the models in producing outputs that better align with users' objectives. Nonetheless, as noted in existing research~\cite{reid-neubig-2022-learning}, the conventional one-time left-to-right generation process of language models contrasts with the iterative refinement and editing approach commonly employed by humans. Furthermore, prior works~\cite{NEURIPS2019_675f9820,reid-zhong-2021-lewis} have demonstrated the efficacy of the generate-and-edit process compared to one-time generation, even with a single editing iteration. Motivated by these findings, this paper explores the integration of large language models (ChatGPT) into an automatic iterative editing pipeline.

% Different from \citet{self_correct}, in which they decompose the generation process into generator and corrector, we instead, decompose it into \textbf{generator, instructor and editor} (see Figure~\ref{fig:pipeline}). It enables us to use large models for the complicated tasks of generation and correction, and employ smaller models to predict user-specific editing instructions.

In contrast to the approach taken by \citet{self_correct}, where the generation process is decomposed into a generator and a corrector, our methodology involves a three-component decomposition consisting of a \textbf{generator, instructor, and editor} (refer to Figure~\ref{fig:intro}). This structure allows us to leverage inference-only large models for the complex tasks of content generation and correction, while utilizing smaller models for the simpler task of generating user-specific editing instructions. The instructor is designed to provide targeted directives for editing and refining the initial outputs of the generator. It is initialized by training on human-authored, or oracle, instructions, which can be obtained by the history of user's behaviour. Following this, the model is then fine-tuned through editor-steered reinforcement learning, wherein the reward function directly quantifies the degree to which\textbf{ the edited output by the editor} align with user preferences, which enhances the model's compatibility with the editor.

% Based on previous works\cite{}, as well as findings of our experiments, ChatGPT is able to serve as both generator and editor, but struggle with generating instructions well aligned with user's needs. \yujia{As this point is mentioned in para 2 already, I think there is no need to bring it up again. This makes the logic confusing, as if you are raising something new.} 
% To effectively generate instructions tailored to the user's preference, we propose to employ a small instructor model, designed to provide targeted directives for editing and refining the initial outputs of the generator. The instructor model is initialized by training on human-authored, or oracle, instructions. Following this, the model is then fine-tuned through editor-steered reinforcement learning, wherein the reward function directly quantifies the degree to which\textbf{ the edited output by the editor} align with user requirements, which enhances the model's compatibility with the editor.

We choose text summarization as the focal task for evaluating this novel framework, which is to generate concise and informative summary for the given document(s). In this paper, we conduct experimental evaluations on two summarization datasets (DeFacto~\cite{liu2022improving} and CNNDM~\cite{cnndm}), focusing on user preference related to factual consistency and coverage. We employ ChatGPT as the generator and the editor model. Our experiments indicate that with the instructions generated by the small instructor model, the edited output is better aligned with user's preference on both datasets. Further experiments on the iterative editing shows that the output can better meet user's needs with more iterations of editing.

\begin{figure*}
    \centering
    \includegraphics[width=\linewidth]{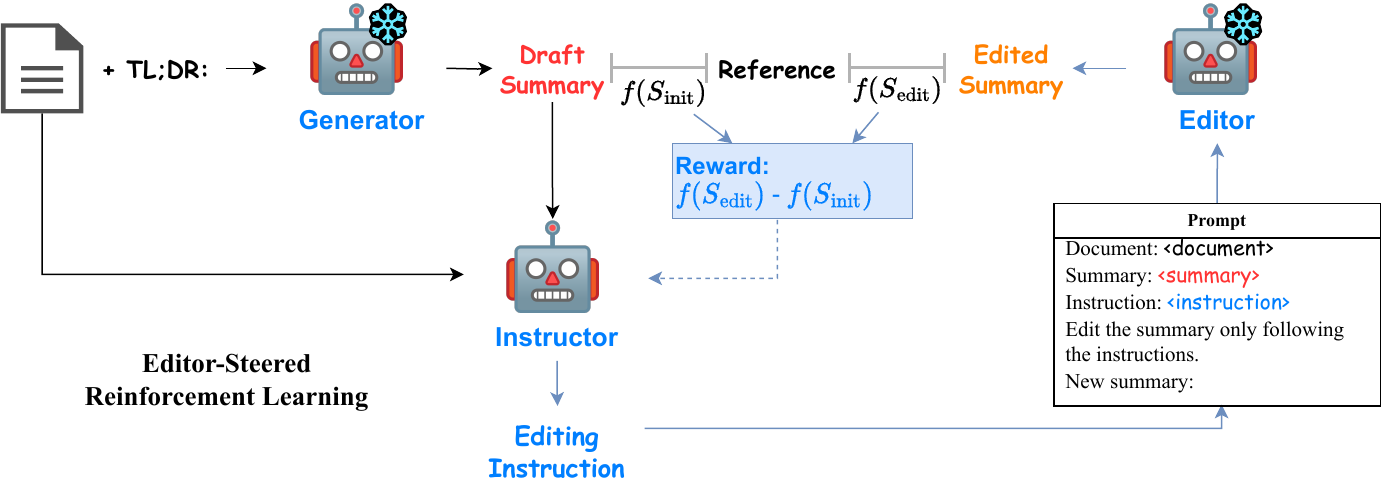}
    \caption{Editor-steered Reinforcement Learning for the instructor. We  fine-tune the instructor using editor-steered reinforcement learning to maximize the expected performance of the editor (e.g., ChatGPT). 
    % \yujia{You may want to update the notations or replace them with natual languages.}\wen{changed.}
    }
    \label{fig:training}

\end{figure*}
\section{Overall Pipeline}
In an effort to enhance the flexibility of the generation pipeline and optimize its compatibility with powerful large language models, we propose a novel decomposition of the generation process into three distinct components, as illustrated in Figure~\ref{fig:pipeline}. These components include: (1) a \textbf{generator}, responsible for producing the initial output; (2) an \textbf{instructor}, tasked with generating natural language instructions that guide the editing of the initial output toward the direction of user preference; 
% \yujia{Will this claim be too strong? As we are doing this in an iterative way, maybe we can do something like ``towards the direction of user requirements''}\wen{sounds good, changed}
and (3) an \textbf{editor}, which refines the initial output in accordance with the provided instructions.

% \wen{new, changed the way to phrase.}
Since it has been demonstrated that large language models can act as both a generator and an editor model, we have chosen to utilize an inference-only large language model, specifically ChatGPT, as our generator and editor. While it is possible to further fine-tune these large language models to serve as instructors, practical limitations such as computational resources~\cite{llama} and access restrictions~\cite{instructgpt} may prevent direct fine-tuning, as has been done in previous works~\cite{self_correct,Second_Thoughts}. Therefore we propose to train a smaller model with editor-steered reinforcement learning to function as a user-specific instructor (as introduced in Section~\ref{sec:pipeline}), which guides the editor in revising the initial output to achieve better alignment with human expectations.

\section{Editor-steered Instructor}
\label{sec:pipeline}

As introduced above, the central objective of the proposed instructor is to produce precise and actionable instructions that can guide a large language model in correcting the original summary to align more closely with the user's preference. To achieve this, we employ a two-phase training process that is designed to enable the instructor to work synergistically with large language models.

% Specifically, given the document $D$, an initial summary is generated using a generator (summarization models or any large language models), i.e. $S_{\text{init}}$, and the goal of the instructor is to take $D$ and $S_{\text{init}}$ as inputs, generating instructions $I=\{i_1,i_2,...,i_k\}$, guiding the editor model to generate edited summaries $S_{\text{edit}}$ better aligned with user requirements.
Specifically, given the document $D$, an initial summary, denoted as $S_{\text{init}}$, is generated using a generator (either a summarization model or a large language model). The objective of the instructor is to take $D$ and $S_{\text{init}}$ as inputs and generate a set of instructions $I = \{i_1, i_2, ..., i_k\}$, aiming to guide the editor model in generating an edited summary that is more closely aligned with the user's preference. Finally, the editor takes $D$, $S_{\text{init}}$, and $I$ as input and generates a revised summary $S_{\text{edit}}$ according to the given instructions.
% \yujia{I think there should be a complete pipeline description \textbf{with notation}, either here or in section 2. Current description is not describing the inputs and outputs of each component clearly.}
% \wen{revised here, does it look good?}
\subsection{Step 1: Supervised Learning}
% In the initial phase of training, we construct a set of oracle instructions based on the user's history for the summary correction.\footnote{The oracle instructions can be simulatively built with the reference summaries from the original training set.}
During the initial training phase, we generate a set of oracle instructions tailored to the user's historical preferences for summary correction.\footnote{These oracle instructions are constructed by simulating the user's preferences using human-written summaries as references, which reflect the distinct summarization preferences of each source. For instance, CNN and DailyMail may exhibit specific tendencies in the summaries it generates for news articles.}
% \yujia{Maybe we can add a footnote about this set is usually converted from the original training set? To avoid possible confusion that our method takes a lot of efforts, even requiring data collection}\wen{added} 
These oracle instructions serve as ideal examples of the instructions that our instructor should produce. We then train the instructor model in a supervised manner, with negative log likelihood loss, i.e.,
$$L=\sum_k P(i_1, i_2,...,i_k|D,S_{\text{init}}).$$ 

The goal of this phase is to establish a solid foundation for the instructor to generate instructions that align with user expectations, by enabling it to learn the relationship between the input (source documents and initial summaries) and the desired output (oracle instructions).

\subsection{Step 2: Editor-steered Reinforcement Learning }
In the second phase, we further fine-tune the instructor model using editor-steered reinforcement learning techniques (see Figure~\ref{fig:training}), specifically using the NLPO algorithm~\cite{ramamurthy2023reinforcement}. 

A key aspect of this phase is the design of the reward function, which serves as the guiding signal for the RL-based fine-tuning process. To ensure that the generated instructions are compatible with the editor model and lead to meaningful summary corrections, the reward function is formulated based on the edited summary, which is generated by the editor model using prompts that include the source documents, initial summaries, and editing instructions provided by the instructor model (see the example prompt shown at right-bottom of Figure~\ref{fig:training}).

To quantify the quality of the edited summary, we employ a scoring function $f(\cdot)$ that measures the extent to which the summary fulfills the user's preference. As we focus on the coverage and factual consistency of the generated summaries as the user's requirements, the scoring function $f(\cdot)$ is then set as the sum of ROUGE score and knowledge coverage, which measures the similarity of the entity level coverage with the reference summaries,
$$f(S) = \alpha \text{ROUGE}(S,S_{\text{ref}})+\beta Cov(S,S_{\text{ref}}).$$
The reward signal itself is defined as the difference in scores between the initial and edited summary, which is designed to capture improvements in summary quality, with higher rewards corresponding to more substantial improvements,
$$Reward = f(S_{\text{edit}})-f(S_{\text{init}}).$$
This phase aims to enhance the model's ability to generate instructions that not only adhere to user requirements, but also effectively guide the large language model to produce improved summaries.

% The quality of the corrected summary is evaluated using a scoring function, which quantifies the extent to which the summary improved regarding the user's requirements.
% By maximizing the reward, the instructor model learns to produce instructions that lead to higher-quality corrected summaries . 

% This phase aims to enhance the model's ability to generate instructions that not only adhere to user requirements but also effectively guide the large language model to produce improved summaries.

% Specifically
% As illustrated in Figure~\ref{fig:training}, the goal of the proposed instructor is to generate precise instructions that can effectively make the large language model correct the original summary towards user's requirements. To make the model works better with the large language models, we train the model with two phases. In the first step, we build the oracle instructions based on the user's requirement, and then train the model in the supervised manner. And in the second step, we further fine-tune the model with reinforcement learning, maximizing the score of the expected summary.

\begin{table*}[t]
    \centering
    \small
    \begin{tabular}{lrrrrr}
    \toprule
    Editor     &DAE&QFE& R1&R2&RL \\
    \midrule
    Initial Summary     &0.699&1.837& 76.03&66.34&74.11\\
   Human Editor&0.906&2.717& 100&100&100\\
    \midrule
    T0PP-D+S+I (Sup)&0.904&2.470&88.74&83.16&87.48
\\
% ChatGPT(zero-shot)&\\
ChatGPT (10-shot)&0.884&2.568&88.48&81.41&86.17\\
    \bottomrule
    \end{tabular}
    \caption{The ROUGE score and factual consistency scores of edited summaries with human-written instructions on DeFacto, in comparison with the human-edited summaries. T0PP-D+S+I (Sup) is a supervised model with the source Documents, initial Summary and Instruction as the input~\cite{liu2022improving}. }
    \label{tab:defacto_cor_summ_gt_instruct}
\end{table*}

\begin{table}[]
    \centering
    \small
    \begin{tabular}{lrrr}
    \toprule
    Model     & R1&R2&RL \\
    \midrule
    ChatGPT (Zero-shot)     & 36.05&22.98&30.66\\
ChatGPT (10-shot)& 37.35&24.94&32.94\\
\midrule
FlanT5 (Sup)& 49.04&34.37&47.07\\
FlanT5 (RL)& 48.05&32.94&46.23\\
% \midrule
% T0 (3B)&51.70&36.56&50.33\\
% T0PP (11B)&52.55&37.41&51.00\\
\bottomrule
    \end{tabular}
    \caption{ROUGE score between generated instructions and human-written instructions on DeFacto. 
    % \hepc{Can we draw a curve for scaling up student model?} 
    }
    \label{tab:defacto_instruction}
\end{table}
\begin{table*}[t]
    \centering
    \small
    \begin{tabular}{lrrrrr}
    \toprule
    Instructor     &DAE&QFE& R1&R2&RL \\
    \midrule
    Initial Summary     & 0.699&1.837&76.03&66.34&74.11\\
    \midrule
    FLAN T5 (Sup) &0.772&2.093& 72.60&61.96&71.21\\
    FLAN T5 (RL)&0.803&2.198& 74.77&64.73&73.44\\

 %    \midrule
 %    Editing (T0PP) 
 % - D+S+I (T0-3B)& 73.01&62.15&-&0.859&2.278\\
    % D+S+I(T0PP)& 75.10&65.15&-&0.859&2.296\\
    % \midrule
    % Editing$_I$ (T5-3B)&77.06&67.86&-&0.804&2.106\\
    % Editing$_I$ (T0-3B)&77.40&68.29&-&0.808&2.112\\
    % Editing$_I$(T0PP)&78.01&69.01&-&0.804&2.108\\
    \midrule
    ChatGPT (10-shot) &0.834&2.583& \textit{56.54}&\textit{41.29}&\textit{53.06}
\\
    \bottomrule
    \end{tabular}
    \caption{The ROUGE score and factual consistency scores of edited summaries with instructions generated by different instructors on DeFacto. We use ChatGPT (10-shot) as the editor model for all the results shown in the table.
    % D+S+I (T0-3B) is a supervised model trained to predict instruction  \citet{liu2022improving}
    }
    \label{tab:defacto_cor_summ}
\end{table*}

\section{Experiments}
We conduct experiments on two distinct datasets, each capturing different facets of user preferences.
\subsection{Scenario 1: Factual Consistency on DeFacto}
% In the first scenario, we choose factual consistency as the user requirements to edit the summary, following \citet{liu2022improving}. To evaluate the effectiveness of our instructor model in addressing factual consistency issues, we utilize the DEFACTO dataset, which is specifically designed for improving factual consistency of the system-generated summaries, with human-annotated demonstrations and feedback. Specifically, for each data example, there is a source document and a system-generated initial summaries, the annotators are asked to write an instruction on how to modify the initial summary to make it more factual consistent, and also write a corrected summary based on the given instructions.
% In the first experimental scenario, we selected factual consistency as the criterion for user's preference to the summary.\footnote{Although factualness might be preferred by most of the users} 
In the initial experimental scenario, we opt to emphasize factual consistency as the primary criterion for users' summary preferences.\footnote{While factual consistency may serve as a typical criterion for summarizers in general, we leverage the instructor to acquire the ability to craft specific instructions that enhance the factual consistency of the summaries.}
% To assess the capability of our instructor model in addressing issues related to factual consistency, 
We employ the DeFacto dataset~\cite{liu2022improving}, a resource specifically curated to enhance the factual consistency of machine-generated summaries through the inclusion of human-annotated demonstrations and feedback. The dataset consists of $701/341/779$ data examples in train/validation/test set respectively.\footnote{Following the original paper, all the experiments are conducted on the examples labeled with errors.} Each data entry in the DeFacto dataset comprises a source document and an initial summary generated by PEGASUS~\cite{pegasus}. Annotators are tasked with providing an instruction that guides the modification of the initial summary to enhance factual consistency. Additionally, annotators generate a revised summary that adheres to the provided instructions and exhibits improved factual consistency.

To evaluate the alignment between the system-generated instructions and the human-written instructions, we employ the ROUGE score as our evaluation metric. Additionally, we assess the quality of the generated summaries with respect to human expectations and factual accuracy using a combination of metrics, including ROUGE scores and factualness scores. Specifically, we utilize the DAE (Dependency Arc Entailment) metric~\cite{goyal2021annotating} and the QFE (Question-answering for Factual Evaluation) metric~\cite{fabbri-etal-2022-qafacteval} to quantify the factualness of the generated summaries. These metrics provide a comprehensive assessment of summary quality in terms of both alignment with human expectations and adherence to factual correctness.
\paragraph{Settings} We use FlanT5-large (700M)~\cite{flant5} as the backbone model for the instructor. The training process for the instructor is executed in two phases, as detailed in Section~\ref{sec:pipeline}. 
% In the initial phase, the instructor is trained using human-authored instructions available in the dataset. Subsequently, in the second phase, editor-steered reinforcement learning is applied to further optimize the instructor, utilizing a reward function defined based on the final edited summaries generated by the editor model. 

\paragraph{Results} 
% We first evaluate the system-generated instructions with the human-written instructions to check whether ChatGPT can identify the user requirements and generate the corresponding instructions, and the results are shown in Table~\ref{tab:defacto_instruction}. It can be found that \textbf{the generated innstructions by ChatGPT can not match the user's requirements well in both zero-shot and few-shot settings}. Then we evaluate whether ChatGPT can be used as a correction model, i.e. correcting the summary according to human's instruction. As shown in Table~\ref{tab:defacto_cor_summ_gt_instruct}, ChatGPT is comparable with the supervised model given the source documents regarding both ROUGE score and factualness scores, initial summaries and editing instructions as inputs, which shows that \textbf{ChatGPT can be used as a summary editor, given the instruction is provided}. In the final experiments (Table~\ref{tab:defacto_cor_summ}), we evaluate whether ChatGPT works well with the system-generated instructions. The results show that the summaries corrected by ChatGPT with 10-shot prompt with the instructions provided by the trained instructor have a much better performance than the previous summary regarding the factualness (DAE/QFE), and the reinforcement learning with the ChatGPT-corporated rewards further boost the improvements. The results are comparable with the supervised corrector with more parameters.

First of all, we assess the potential of ChatGPT to serve as an editor model, capable of revising summaries in accordance with human-provided instructions. The results of this assessment, presented in Table~\ref{tab:defacto_cor_summ_gt_instruct}, indicate that ChatGPT performs comparably to a supervised model when supplied with source documents, initial summaries, and human-written editing instructions as input, as demonstrated by comparable ROUGE scores and factualness scores. These findings affirm that ChatGPT is effective as a summary editor when appropriate editing instructions are provided.

Then, we evaluate the system-generated instructions in comparison to human-authored instructions. Our objective is to determine the extent to which ChatGPT and trained instructors can accurately discern user requirements and subsequently produce corresponding instructions. The results of this evaluation are presented in Table~\ref{tab:defacto_instruction}. Notably, we observe that the instructions generated by ChatGPT do not effectively match human-written instructions, as evidenced by suboptimal performance in both zero-shot and few-shot settings. Although the instructor model we used is much smaller than ChatGPT (700M v.s. 175B), it shows the ability to generate instructions better aligned with the user's needs. 

In the final set of experiments, presented in Table~\ref{tab:defacto_cor_summ}, we evaluate the performance of the editing model (ChatGPT) with the trained and RL fine-tuned instructors, as well as the instructions generated by ChatGPT in few-shot settings. The results demonstrate that summaries edited by ChatGPT, when utilizing a 10-shot prompt and instructions from the trained instructor, exhibit large improvements in factualness(as measured by DAE/QFE) compared to the original summaries . The implementation of reinforcement learning, incorporating ChatGPT-derived rewards, leads to additional enhancements in summary quality. 
% Impressively, the performance of ChatGPT in this context is comparable to that of a supervised corrector model with a larger number of parameters, underscoring the effectiveness of ChatGPT in summary correction and editing tasks. 
Furthermore, we conduct experiments utilizing instructions generated by ChatGPT. While these instructions demonstrate suboptimal alignment with human-authored instructions, they yield unexpectedly high scores in terms of factualness, particularly as measured by the QFE metric. However, a notable decrease in ROUGE scores is observed in comparison to other methods. These findings suggest that ChatGPT possesses the capacity to generate instructions that target a specific and well-defined aspect (e.g., addressing factual inconsistencies), but may struggle to accurately discern and fulfill broader human expectations.

\subsection{Scenario 2: Coverage on CNNDM}

\begin{table}[]
    \centering
    \small
    \begin{tabular}{lrrrr}
    \toprule
    Instructor     & Knlg F1 & R1&R2&RL\\
    \midrule
    Initial Summary  &44.15&40.28&16.65&33.23\\
    \midrule
    FLAN T5 (Sup)&47.44&41.04&16.72&33.63\\
    FLAN T5 (RL)&47.99&41.21&16.80&33.90\\
    ChatGPT (5-shot)* &43.43&39.46&15.43&32.40\\
    \midrule
   Oracle&60.80&43.08&18.37&35.24\\
    \bottomrule
    \end{tabular}
    \caption{Knowledge coverage  and ROUGE scores of edited summaries with instructions generated by different instructors on CNNDM. We use ChatGPT (zero-shot) as the generator model (to produce Initial Summary) and editor model. * We reduce the number of examples in the prompt if it exceeds the length limit (4k tokens).}
    \label{tab:cnndm_results}
\end{table}

ChatGPT has demonstrated its capacity to produce fluent and informative summaries of news articles~\cite{goyal2022news}. Despite its proficiency in generating coherent summaries, it may not always achieve the desired coverage of key topics, as expected by the user. In response to this challenge, we conduct an experiment to train and evaluate an instructor model specifically designed to guide the editing of summaries for improved knowledge coverage based on user's history. The instructor predicts the keywords to be added to or removed from the current summary, thereby providing actionable instructions to align the summary more closely with user preference. In practice, we assess knowledge coverage based on the extent to which the generated summaries match reference summaries in terms of keyword content.

We employ the CNNDM dataset~\cite{cnndm} as our benchmark for this experiment, which contains pairs of articles and reference summaries, with the original reference summary serving as the target representation of user preference on the coverage. We acknowledge that, according to recent studies~\cite{goyal2022news}, the reference summaries in the CNNDM dataset may exhibit some quality limitations, such as poor coherence. However, our primary focus in this experiment is on knowledge coverage rather than summary quality. We are interested in assessing the extent to which the generated summaries capture the key entities in the reference.

To measure knowledge coverage,
% \hepc{Should we have a formula to describe how we calcuate KF1?}
we introduce an entity-level matching metric Knlg F1.
Let $E_{\text{gen}}$ be the entities mentioned in the generated summaries and $E_{\text{ref}}$ be those in the reference summaries. We quantify the degree of overlap between the two by
\begin{eqnarray*}
    \text{Knlg F1}= \frac{2 \text{Knlg}_\text{p}\times \text{Knlg}_\text{r}}{\text{Knlg}_\text{p}+\text{Knlg}_\text{r}}, ~\text{where}\qquad\qquad~~~ \\
    \text{Knlg}_\text{p}=\frac{|E_{\text{ref}}\cap E_{\text{gen}}|}{|E_{\text{gen}}|}, ~~
    \text{Knlg}_\text{r}=\frac{|E_{\text{ref}}\cap E_{\text{gen}}|}{|E_{\text{ref}}|}.
\end{eqnarray*}
By maximizing this overlap, the instructor aims to produce summaries that effectively cover pertinent information as indicated by the reference.

\paragraph{Settings:} We use the summaries generated by ChatGPT as the initial summaries.\footnote{The dataset is released, and can be found in the Github repo.}. And we employ FlanT5-large (700M) as the instructor model for predicting keywords, using both the original document and the initial summaries generated by ChatGPT as input. Supervised training is performed using oracle keyword lists specifying which keywords to add and remove. Subsequently, the model undergoes editor-steered reinforcement learning fine-tuning, as detailed in Section~\ref{sec:pipeline}, using a subset of 10,000 training examples from the dataset for efficiency.
% We use FlanT5-large (700M) to predict the keywords, with the original document and the initial summaries as input. The initial summaries are generated by ChatGPT itself. We train the model with the oracle keyword lists (add and remove) in the supervised settings, and further train the model with reinforcement learning as described in Section~\ref{sec:pipeline} on a subset with 10k training examples.
\begin{table}[]
    \centering
    \small
    \begin{tabular}{lrrrr}
    \toprule
    Model     & Knlg F1 & R1&R2&RL\\
    \midrule
    Initial Summary     &44.15&40.28&16.65&33.23\\
    \midrule
    Edit Iter 1&47.99&41.21&16.80&33.90\\
    Edit Iter 2&48.65&41.18&16.69&33.88\\
    Edit Iter 3&48.99&41.14&16.63&33.83\\
    \midrule
    Edit Iter 1 (1\&2)&48.08&41.25&16.91&33.94\\
    Edit Iter 2 (1\&2)&48.87&40.62&16.60&33.45\\
    Edit Iter 3 (1\&2)&49.20&41.15&16.87&33.86\\
    \bottomrule
    \end{tabular}
    \caption{Iterative editing on CNNDM. The second block shows the results of the model fine-tuned on the data in the first iteration only, and the bottom block shows that of the model fine-tuned on the data in the first and second iterations.}
    \label{tab:iter_editing}
\end{table}
\begin{table*}[]
    \centering
    \small
    \begin{tabular}{lp{0.7\linewidth}}
    \toprule
    Initial Summary&A former corrections officer was punched by a young man on a plane after he asked him to stop using foul language. The former officer then took the young man down and held him until police arrived. Source: \textcolor{red}{Daily Mail} \\
    Oracle Instruction& <Add> \textcolor{Green}{Chad Hurst} <remove> \textcolor{red}{Daily Mail}\\
    Human-written Reference&\textcolor{Green}{Chad Hurst} of Salt Lake City, Utah was sucker punched by a plane passenger when they landed in the city Sunday . This after Hurst asked the young man to stop using foul language following their flight . Hurst, a former corrections officer, then took down the man and pinned his arms behind his back while waiting for law enforcement . The young man, who has still not been named by police, was charged with assault and public intoxication .\\
    Predicted Instruction&  <Add> \textcolor{Green}{Chad Hurst} <remove> \textcolor{red}{Daily Mail}\\
    ChatGPT-edited Summary&  \textcolor{Green}{Chad Hurst}, a former corrections officer from Salt Lake City, Utah, was punched by a young man on a plane after he asked him to stop using foul language. Hurst calmly took the young man down and held him until police arrived. The young man was charged with assault and public intoxication. Hurst's training as a former corrections officer taught him to never punch back but to control the situation and take the person down. \\
    \bottomrule
    \end{tabular}
    \caption{An example from the CNNDM dataset.}
    \label{tab:examples_cnndm}
\end{table*}
\paragraph{Results:} 
% The experiment results with the knowledge coverage and ROUGE scores are shown in Table~\ref{tab:cnndm_results}. ChatGPT has shown a good performance as a summarizer in the zero-shot setting, and meanwhile, with the oracle instruction encoded, it also shows the ability to correct the initial summary according to the given instruction. By using the instructions generated by the trained model, both ROUGE scores and knowledge coverage have improved, and the reinforcement learning further boost the performance wihtin a moderate range. However, when directly using ChatGPT to generate the instruction in the few-shot setting, the corrected summaries show a even lower score regarding both knowledge coverage and ROUGE scores than the initial summary.
The results of our experiments, presented in Table~\ref{tab:cnndm_results}, demonstrate the effectiveness of our instructor model in enhancing knowledge coverage, indicated by both entity matching and ROUGE scores. In a zero-shot setting, ChatGPT exhibits strong performance as a summarizer. Importantly, when provided with Oracle instructions, ChatGPT also demonstrates a robust capacity to correct and refine initial summaries in accordance with the specified instructions.

The integration of instructions generated by our trained instructor model leads to remarkable improvements in knowledge coverage, indicating that the summaries better align with user preference (comparing FLAN T5 (Sup) with Initial Summary). Moreover, we observe that the reinforcement learning fine-tuning process(FLAN T5 (RL))further improves the model's performance, resulting in moderate but meaningful gains in the evaluated metrics.

In contrast, when we explore a few-shot setting in which ChatGPT directly generates instructions without the use of the trained instructor(ChatGPT (5-shot)), the edited summaries exhibit a decline in performance. Specifically, both Knlg F1 and ROUGE scores are lower than those of the initial summaries, suggesting limitations in ChatGPT's ability to generate effective instructions for summary editing to better align with users' preference.

Overall, these findings underscore the value of our instructor as a powerful intermediary for guiding large language models such as ChatGPT in editing summaries to more closely adhere to user preference. 
% The ability to enhance knowledge coverage and improve summary quality through instruction-based guidance represents a compelling advancement in the field of natural language summarization.

\section{Discussion}
\subsection{Iterative Editing}
In addition to performing one-step editing, we conducted experiments to explore the effectiveness of iterative editing on the CNNDM dataset\footnote{We did not conduct similar experiments on the DeFacto dataset because, for the majority of data examples, only one editing step is required to transition from the initial summary to the human-edited summary}. The results of the iterative editing experiments are presented in Table~\ref{tab:iter_editing}. Utilizing reinforcement learning (RL) training based solely on data from the first iteration, we observed an improvement in the coverage of the edited summaries over the iterative editing process.
% ; however, coverage declined in the third iteration \hepc{I didn't see the decline. Do you mean ROUGE? }. A possible explanation for this decline is that the data distribution underwent significant changes after two rounds of editing. To address this, 
We further fine-tuned the model using a mixture of data from both the first and second iterations, which leads to improved performance, as evidenced by enhanced knowledge F1 in the iteratively edited summaries.
% In addition to the one-step editing, we also run an experiments on iterative editing on the CNNDM dataset\footnote{We do not do the same experiments on DeFacTO because for most of the data examples, there is only one editing step from the initial summary to the human edited summary}. The results are shown in Table~\ref{tab:iter_editing}. Trained With RL on the data in the first iteration only, the coverage of the edited summaries improves in the first two iterations, however, it decrease in the third iterations. One possible reason is that the data distribution has significantly changed after two iterations o edition. Thus we further fine-tune the model with the mixed data from the first and second iteration. As shown below, after fine-tuning with the data from the second iteration, the keyword matching shows a better performance. 

\subsection{Qualitative Examples}

We show examples from the CNNDM dataset in Table~\ref{tab:examples_cnndm}. The instructor model can correctly detect the user's expectation and produce the editing instruction. ChatGPT is capable to edit the initial summary based on the given instruction, serving as an editor. \footnote{Examples from DeFacto  are shown in the appendix.}
\section{Related Work}
\subsection{Text Editing}
% Post-editing techniques have been extensively investigated in various NLP tasks, including sentence fusion~\cite{malmi-etal-2019-encode}, style transfer~\cite{reid-zhong-2021-lewis}, and wiki-editing~\cite{reid-neubig-2022-learning,faltings-etal-2021-text}. These methods often entail editing through a series of micro-defined operations, such as insertion, deletion, and replacement. While effective, these methods typically necessitate a substantial volume of human-labeled data or the construction of intricate editing chains. In contrast to these approaches, our work explores text editing at a more abstract level, utilizing natural language editing instructions.  This higher-level approach enables us to leverage the capabilities of large language models, such as ChatGPT. Similarly, \citet{liu2022improving} proposed an approach involving a critic model for feedback generation and an editor model for revising initial summaries. We extend this approach by formalizing it as an iterative editing pipeline and further enhance it by leveraging inference-only large language models within the pipeline and training an editor-steered instructor.
% \citet{Second_Thoughts} introduced a novel training paradigm for language models that involves re-aligning generated text with human values through a dynamic programming (DP)-derived chain-of-edits. However, this method entails additional fine-tuning of the language model, which may not be feasible for models with limited resources and accessibility.
Post-editing techniques have been extensively studied in various NLP tasks, including sentence fusion \cite{malmi-etal-2019-encode}, style transfer \cite{reid-zhong-2021-lewis}, and wiki-editing \cite{reid-neubig-2022-learning,faltings-etal-2021-text}. These methods involve micro-defined operations such as insertion, deletion, and replacement. However, they often require a substantial amount of human-labeled data or complex editing chains. In contrast, our work focuses on abstract-level text editing using natural language instructions, leveraging the capabilities of large language models like ChatGPT. Similarly, \citet{liu2022improving} propose an approach involving a critic model for feedback generation and an editor model for revising initial summaries. We extend this approach by formalizing it as an iterative editing pipeline and enhancing it with inference-only language models and an editor-steered instructor.

Recently, \cite{Second_Thoughts} introduced a novel training paradigm that aligns generated text with human values through a dynamic programming-derived chain-of-edits. However, this method requires additional fine-tuning of the language model, which may be impractical for models with limited resources and accessibility.
% We expand the approach by formalizing the method as an iterative editing pipeline, and further enhance by employ the inference-only large language models in the pipeline  and train a editor-steered instructor.

% (e.g., open-source models like LLAMA~\cite{llama}) or restricted access (e.g., API-based models such as ChatGPT).

In another line of work, \citet{self_correct} proposed a framework that decomposes the original generation process into generator and corrector components, where the corrector is trained through online training to iteratively refine imperfect generations. Our work differs from them by decomposing the generation process into three components: the generator, the instructor, and the editor. This decomposition allows us to utilize large models for complex generation and correction tasks, while employing smaller models to predict user-specific editing instructions.

In parallel to our research, \citet{madaan2023selfrefine} propose a similar generation pipeline aimed at iteratively refining the generated output. However, their approach differs in that they utilize the same large language model (with varying prompts) for generating the initial output, providing feedback, and editing the output based on the received feedback, without considering any user-specific feedback. In contrast, our focus in this paper is on aligning the generated output more closely with user needs, guided by a trained instructor.
% \paragraph{Text Editing: }  Post-editing methods have been explored widely in sentence fusion\cite{malmi-etal-2019-encode}, style transfer\cite{reid-zhong-2021-lewis} and WIKI-editing\cite{reid-neubig-2022-learning,faltings-etal-2021-text}. However, most of them perform editing with micro-defined steps, e.g. insertion, deletion, replacement, and etc., which either require large amount of human-labeled data, or a carefully-built editing chain. Different from the previous works, we explores the editing process on a higher level, with natural language editing instructions, which can take benefits from the powerful large language model (e.g. ChatGPT). \citet{Second_Thoughts} propose a new training paradigm for language model by re-align the generated text with human value though a DP-derived chain-of-edits, however, it requires additional fine-tuning on the language model, which may not be feasible for limited resource(open-source models, like LLAMA\cite{llama}) and access (API-based models, e.g. InstructGPT\cite{instructgpt} and ChatGPT). \citet{self_correct} decompose the original generation process into generator and corrector, the latter is then trained through  online training procedure, learning to iteratively correct imperfect generations. Different from their work, we instead, decompose the generation process into generator, instructor and editor, by which we can use large models for the complicated tasks of generation and correction, and employ smaller models to predict user-specific editing instructions.
\subsection{Large Language Models}
The field of natural language processing has witnessed significant advancements in the realm of large language models (LLMs)~\cite{palm,opt,lambda}, leading to the creation of models that exhibit extraordinary language processing capabilities. Among these models, the GPT family~\cite{gpt3} stands as a prominent example, earning widespread recognition for its versatile performance across different language-related tasks. 
% Additional noteworthy language models include PaLM~\cite{palm}, and OPT~\cite{opt}.

The introduction of instruction tuning~\cite{instruct_tuning} has further catalyzed the enhancement of language models, particularly when trained with human instructions~\cite{t0}. Notably, this approach has resulted in substantial improvements, especially within the context of zero-shot and few-shot learning. InstructGPT~\cite{instructgpt}, which employs the Reinforcement Learning from Human Feedback (RLHF) training paradigm, exemplifies this trend, enabling models to effectively follow human instructions and providing a foundational basis for our current work.

The recent release of  LLAMA~\cite{llama} has further expanded opportunities for exploration in this area, as researchers have begun to train or fine-tune models using task-augmented datasets by GPT models~\cite{self_instruct}.

Distinct from the aforementioned research efforts, our work introduces the tri-agent pipeline, a novel paradigm that capitalizes on the capabilities of large language models for downstream tasks. Uniquely, our approach is designed to optimize performance while minimizing computational resource demands and accommodating limited access to large language models (e.g., API-only access).

% With the proposal of instruction tuning\cite{}, the language models trained with human instructions shows a large improvements\cite{}, especially in zero-/few-shot settings. With the RLHF training paradigm proposed in InstructGPT, the models can behave following human's instructions, which provide the basis for this work. Recently, with the open source large language model llama released, more works started to train the models with instructions or fine-tune them on task specific datasets augmented by the GPT models. Different from all the work above, we propose the generate-instruct-edit pipeline, as a novel paradigm on making use of the large language models in the downstream tasks, with low computational resource and limited access to the large language model (API only).
\subsection{Summarization with LLM}
Before the advent of LLMs, a prevalent approach to the text summarization task involved pre-training models on a substantial corpus using task-focused objectives, followed by fine-tuning on task-specific datasets. This paradigm demonstrated effectiveness in text summarization and was adopted by models such as PEGASUS~\cite{pegasus}, Primera~\cite{primera}, and Z-Code++~\cite{he2022zcode}. However, recent studies~\cite{goyal2022news,zhang2023benchmarking} have revealed that the application of GPT-3~\cite{gpt3} and InstructGPT~\cite{instructgpt} to news summarization tasks in zero-shot settings yields results that are not only preferred by human evaluators over those of supervised models, but are also more favorable than the reference summaries themselves.

These findings suggest a  direction for the text summarization task. Rather than training supervised summarizers on potentially suboptimal reference summaries, it may be more efficient to leverage LLMs, and focus on editing their outputs to align with user requirements, which is also in-line with the tri-agent pipeline proposed in this work.
% This shift in perspective underscores the potential for large language models to play a pivotal role in advancing the field of text summarization.

% Prior to GPT3\cite{gpt3} being introduced, it has been shown that pre-training on a large corpus with task-focused objective then finetuning on task-specific specific dataset is an effective paradigm on the text summarization task~\cite{pegasus,he2022zcode,primera}. However, recent works~\cite{goyal2022news,zhang2023benchmarking}  find that applying GPT3 and InstructGPT to the summarization task in zero-shot settings is more preferrable than the supervised models by the human evaluators, and it's even more favorable than the reference summaries. It sheds light on the future direction of the summarization task, instead of training supervised summarizers on those suboptimal references, it may be more effecient to edit the output of the large language models towards user's requirements.

% \begin{itemize}
%     \item Editing model
%     \item chatgpt
%     \item rl with gpt reward?
% \end{itemize}
\section{Conclusion and Future Work}
In this paper, we introduce a novel generation paradigm that decomposes the generation process into three distinct components: the generator, the instructor, and the editor. Our approach is specifically designed to harness the capabilities of large language models, while accounting for constraints such as limited access and computational resources, and to facilitate the customization of generated content to align with user preference. Through a series of pilot experiments on the task of text summarization, we find that large language models, exemplified by ChatGPT, can effectively serve as editors, achieving performance levels comparable to supervised editing models when provided with human-written instructions. Nevertheless, it is still challenging for the large language models to generate instructions that are well-aligned with human-authored instructions.

To address this challenge, we employ a smaller model as the instructor, which is trained
% in a supervised setting using user-specific human-authored (oracle) instructions and further fine-tuned using 
with editor-steered reinforcement learning (RL) with rewards based on the quality of the edited summaries. Our experimental results demonstrate the efficacy of this approach in guiding the editor (ChatGPT) to produce summaries that are more closely aligned with user expectations.

Looking ahead, future work will involve extending our experiments to other tasks, such as wiki-editing~\cite{reid-neubig-2022-learning}, news-editing~\cite{spangher2022newsedits}, and mathematical problem synthesis~\cite{self_correct}. Additionally, we may generate more instruction data using the self-instruct technique~\cite{self_instruct} to train a better instructor.

\section*{Limitations}
While our proposed generation pipeline aims to improve the alignment of large language model outputs with user preference, we acknowledge the limitation of resource constraints in our study. As a result, we focus our experiments solely on ChatGPT, which has demonstrated top performance across a range of tasks. However, future work should explore its applicability and performance with other large language models as well. Furthermore, it is important to note that, like all large language models, our system's output may still exhibit issues such as hallucination and bias. While our pipeline partially addresses these concerns, we cannot guarantee that the results are completely free from hallucination and bias. 
 
% In this paper, we propose a new generation paradigm, decomposing the process into three components, generator, instructor and editor. It is designed to effectively make use of the large language models with limited access and computational resource, while making the generation customized to meet user's requirements. In the pilot experiments on the summarization task, we find that the large language models (ChatGPT) is able to act as the editor, and it has a comparable performance with the supervised editing models with the human instructions. However, it fails to generate instructions well aligned with human written instructions. Therefore we employ a smaller model serving as the instructor, trained on the user-specific human-written/oracle instructions in a supervised setting, followed with RL directly on the quality of edited summaries. The experiments show that our method can effectively guide the editor(ChatGPT) to generate summaries better aligned with user's expectations. 

% For future work, we will run experiments on other tasks or domains, like WIKI-editing~\cite{reid-neubig-2022-learning}, news-editing~\cite{spangher2022newsedits}, or mathematical problem synthesizer~\cite{self_correct}. In addition, 
% \section*{Limitations}

% \section*{Ethics Statement}

% \section*{Acknowledgements}

\bibliography{anthology,custom}
\bibliographystyle{acl_natbib}
\appendix

\section{Prompts}
\label{sec:all_prompts}
We show the prompts used for summary editing and  instruction generation in Table~\ref{tab:prompt_summary} and Table~\ref{tab:prompt_instruction}, respectively. 
\begin{table}[h!]
    \small
    \centering
    \begin{tabular}{p{0.9\linewidth}}
    \toprule
    CNNDM\\
    \midrule
    Summary: [initial summary] \\
    Document: [article] \\
    Rewrite the summary for the document, [instruction]\\
    New summary: \\
    \midrule
    DeFacto\\
    \midrule
    Document: [article]\\
    Summary: [initial summary] \\
    Instructions: [instruction] \\
    Edit the summary only following the instructions and only output the corrected summary.\\
    New summary: \\
    \bottomrule
    \end{tabular}
    \caption{Prompts used for summary editing.}
    \label{tab:prompt_summary}
\end{table}

\begin{table}[]
    \small
    \centering
    \begin{tabular}{p{0.9\linewidth}}
    \toprule
    CNNDM\\
    \midrule
    \textit{few-shot prompts }$\times N$\textit{, up to the length limit}\\
    Document: [article]$_i$ \\
    Summary: [initial summary]$_i$\\
    Instructions: [instruction]$_i$\\
    \\
    Document: [article]\\
    Summary: [initial summary]\\
    The summary may not cover the salient content, generate instructions to make the summary focus on salient content. 
    The instructions should be chosen from the following formats: \\
    Delete content related to \_\_. \\
    Add content related to \_\_. \\
    No operation is needed.\\
    Only output the instructions without the corrected summaries, and make the instruction conservatively. \\
    Instructions: \\
    \midrule
    DeFacto\\
    \midrule
    \textit{few-shot prompts }$\times 10$\\
    Document: [article]$_i$\\
    Summary: [initial summary]$_i$ \\
    The summary may contain some factual errors, generate the instructions to correct the summary. \\
    Instructions: \\
        \\    
    Document: [article] \\
    Summary: [initial summary] \\
    The summary may contain some factual errors, generate the instructions to correct the summary. \\
    The instructions should be chosen from the following formats: \\
    Remove the information about \_\_ from the summary. \\
    Add the information about \_\_ to the summary. \\
    Replace the information about \_\_ with the in-formation about \_\_. \\
    Modify the information about \_\_ in the summary. \\
    Rewrite the summary entirely by \_\_. \\
    Only output the instructions without the corrected summaries, and make the instruction conservatively. \\
    Instructions: \\
    \bottomrule
    \end{tabular}
    \caption{Prompts used for instruction generation}
    \label{tab:prompt_instruction}
\end{table}
\section{Qualitative Examples}
We show examples from the DeFacto dataset in Table~\ref{tab:examples_cnndm}. The instructor model can correctly detect the user's expectation and produce the editing instruction. ChatGPT is capable to edit the initial summary based on the given instruction, serving as an editor.
\begin{table*}[]
    \centering
    \small
    \begin{tabular}{lp{0.7\linewidth}}
    \toprule
    Initial Summary&The controversial Kudankalum nuclear power station in \textcolor{red}{India's Tamil Nadu state} has started generating electricity. \\
    Human-written Instruction&Remove the information about \textcolor{red}{the location of India's Tamil Nadu state} from the summary.\\
    Human-edited Summary& The controversial Kudankalum nuclear power station has started generating electricity.\\
    Predicted Instruction& Remove the information about \textcolor{red}{Tamil Nadu} from the summary.\\
    ChatGPT-edited Summary& The controversial Kudankalum nuclear power station has started generating electricity.\\
    \midrule
    Initial Summary& Gunfire has been heard in Ivory Coast's \textcolor{red}{second} city of Bouaké, a day after soldiers mutinied over pay\\
    Human-written Instruction&Remove the information about \textcolor{red}{second} from the summary.\\
    Human-edited Summary&Gunfire has been heard in Ivory Coast city of Bouaké, a day after soldiers mutinied over pay.\\
    Predicted Instruction&Remove the information about \textcolor{red}{second} from the summary. \\
    ChatGPT-edited Summary&Gunfire has been heard in Ivory Coast's city of Bouak, a day after soldiers mutinied over pay.\\
    \bottomrule
    \end{tabular}
    \caption{Examples from the DeFacto dataset.}
    \label{tab:examples}
\end{table*}
\section{Software and Licenses}
Our code is licensed under Apache License 2.0. Our framework dependencies are: 
\begin{itemize}
    \item HuggingFace Datasets\footnote{\url{https://github.com/huggingface/datasets/blob/master/LICENSE}}, Apache 2.0
    \item NLTK \footnote{\url{https://github.com/nltk/nltk}}, Apache 2.0
    \item Numpy\footnote{\url{https://github.com/numpy/numpy/blob/main/LICENSE.txt}}, BSD 3-Clause "New" or "Revised"
    \item Transformers\footnote{\url{https://github.com/huggingface/transformers/blob/master/LICENSE}}, Apache 2.0
    \item Pytorch\footnote{\url{https://github.com/pytorch/pytorch/blob/master/LICENSE}}, Misc
    \item ROUGE \footnote{\url{https://github.com/google-research/google-research/tree/master/rouge}}, Apache 2.0
    \item Flan T5
    \footnote{\url{https://huggingface.co/google/flan-t5-large}},
    Apache 2.0
    \item ChatGPT
    \footnote{\url{https://openai.com/chatgpt}},
    Proprietary
\end{itemize}
\end{document}